\documentclass{article}

\usepackage[T1]{fontenc}
\usepackage[english]{babel}

\usepackage[a4paper,
            top=2cm,
            bottom=2cm,
            left=3cm,
            right=3cm]{geometry}

\usepackage{graphicx}

\usepackage{subcaption}
\usepackage{booktabs}
\usepackage{xcolor}
\usepackage{tcolorbox}

\usepackage{amsmath}
\usepackage{amssymb}

\usepackage[disable]{todonotes}

\usepackage{siunitx}

\usepackage{float}
\usepackage{caption}

\usepackage{algorithm}
\usepackage{algpseudocode}

\usepackage{authblk}

\usepackage[colorlinks=true, allcolors=blue]{hyperref}

\usepackage{amsthm}
\theoremstyle{remark}
\newtheorem{remark}{Remark}
\usepackage{enumitem}

\usepackage[numbers]{natbib}

\newcommand{\neu}{node}
\newcommand{\DT}{\textit{technical definition}}
\newcommand{\CDC}{\textit{Requirements Specification}}

\newcommand{\ASUPPRIMER}[1]{\textcolor{red}{Several lines deleted here}}

\title{CRADIPOR: Crash Dispersion Predictor}

\author[1,2]{Edgar Chaillou}
\author[1]{Sebastian Rodriguez}
\author[1]{Yves Tourbier}
\author[1,3]{Francisco Chinesta}

\affil[1]{PIMM, Arts et M\'etiers Institute of Technology, Paris, France}
\affil[2]{Renault Group, Guyancourt, France}
\affil[3]{ENSAM Institute of Technology, CNRS@CREATE, Singapore}

\begin{document}

\maketitle

\begin{abstract}
We present CRADIPOR, a numerical dispersion prediction tool for automotive crash simulations. Finite Element (FE) crash models are widely used throughout vehicle development, but their predictions are not strictly repeatable because of parallel computation and model complexity. As a result, performance criteria evaluated during post-processing may exhibit significant numerical dispersion, which complicates engineering decision-making. Although dispersion can be estimated by repeating the same simulation, this approach is generally impractical because of its high computational cost.

This work therefore investigates a prediction tool that can be applied during routine crash-simulation post-processing without repeating the computation. The proposed approach relies on a Rank Reduction Autoencoder (RRAE) combined with supervised classification in order to identify regions sensitive to numerical dispersion. The comparative analysis suggests that the RRAE-based framework is more effective than the Random Forest baseline on the studied dataset. Among the tested signal representations, wavelet-based and slope-based inputs appear to be the most promising, with slope variations providing the best classification performance. These results support the use of structured latent representations for improving numerical-dispersion detection in automotive crash post-processing.
\end{abstract}




\newpage

\section{Introduction}
\label{sec:problem}

\begin{figure}[ht]
    \centering
    
    \begin{subfigure}[t]{0.48\linewidth}
        \centering
        \includegraphics[width=\linewidth]{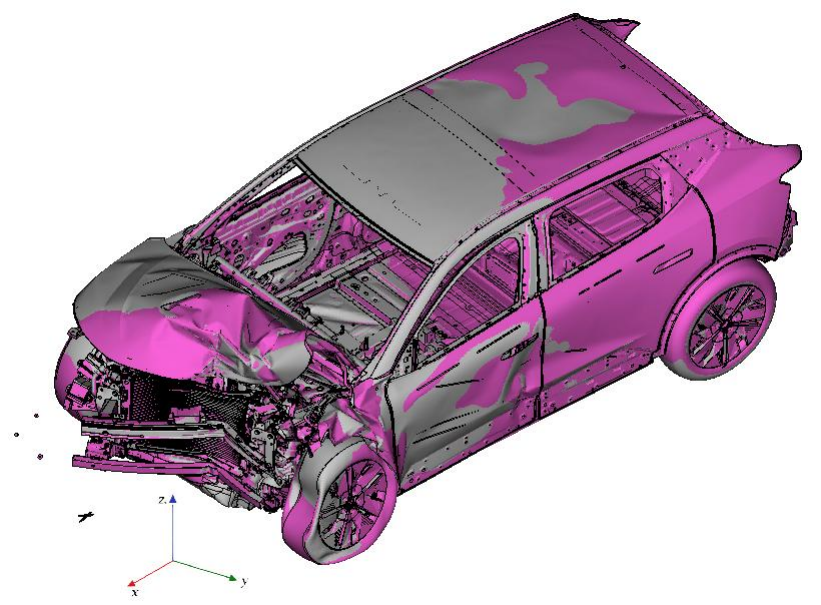}
        \caption{Superposition of two nominally identical simulations at timestep 220.}
        \label{fig:initial_model_crashed}
    \end{subfigure}
    \hfill
    \begin{subfigure}[t]{0.48\linewidth}
        \centering
        \includegraphics[width=\linewidth]{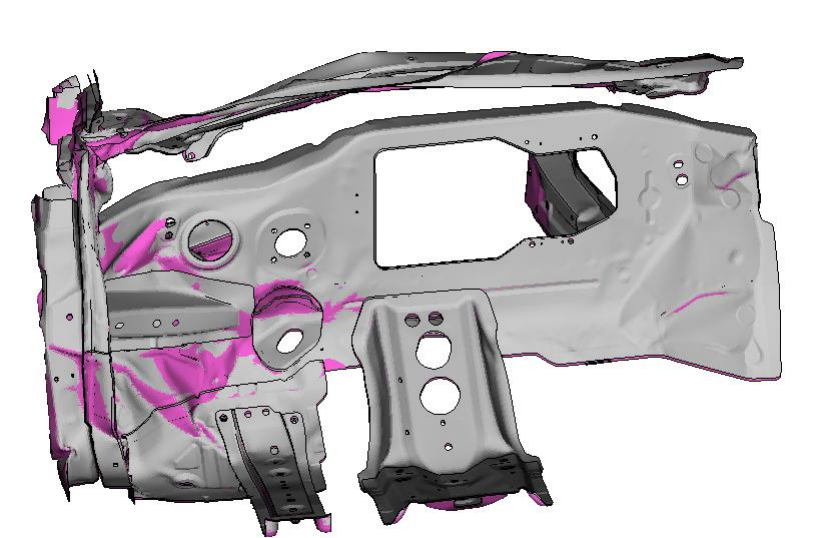}
        \caption{Zoom on the footwell region, associated with the evaluation of lower limb injury criteria.}
        \label{fig:dispersion_illustration}
    \end{subfigure}
    
    \caption{Illustration of numerical dispersion: global comparison (left) and local detail (right).}
    \label{fig:dispersion_global_local}
    
    \end{figure}

    Design engineers generally use deterministic simulations to ensure reproducibility. A simulation is intended to reproduce a defined case. The consideration of manufacturing tolerances and other sources of variability is generally addressed either through a nominal design approach with safety margins or by means of robust design methods (FORM / SORM, etc.)~\cite{rackwitz1978structural,breitung1984asymptotic}. Crash simulation constitutes a particular case: parallel computation and explicit time integration schemes inherently generate dispersion~\cite{bathe2006fem,collange2015numerical,belytschko2013nonlinear}. Industrial crash models are assemblies of various numerical objects: steel structures meshed with shell elements, cast parts meshed with volumetric finite elements, mesh-free airbags, aluminum tube crash barriers, dummies, sensors, foams, fuel, etc. These models represent a compromise between the quality of physical representation, in particular, mesh size, and computational cost~\cite{bathe2006fem,belytschko2013nonlinear}. It is therefore observed that two simulations performed with strictly identical parameters may lead to different final responses (illustrated in \autoref{fig:dispersion_global_local}). It is possible to make the simulation nearly reproducible, but at a very high computational cost, without any guarantee that the result is more accurate.
    
    Numerical dispersion does not represent manufacturing variability, but it may instead reveal physical phenomena of direct interest to the designer, such as buckling~\cite{timoshenko1961elastic}. Real crash events are inherently unstable due to the strong nonlinearities associated with large deformations, multiple contacts, material failure, and irreversible plastic behavior~\cite{bathe2006fem,belytschko2013nonlinear}. The simultaneous interaction of these mechanisms makes the global response particularly sensitive to initial conditions, a characteristic commonly observed in nonlinear dynamic systems~\cite{strogatz2018nonlinear}. Furthermore, modeling choices contribute additional sources of sensitivity.
            
    This sensitivity is further amplified by the dynamic nature of crash simulations. Explicit time integration schemes require very small time steps to ensure numerical stability~\cite{bathe2006fem}. In this context, minor numerical perturbations arising from machine rounding errors, spatial or temporal discretization, or contact algorithms may be amplified during computation and lead to divergences between nominally identical runs~\cite{goldberg1991floating,higham2002accuracy}.
        
    Several parameters directly influence the magnitude of this dispersion, including mesh size and quality, the selection of material constitutive laws, contact interface modeling, and finite element formulations~\cite{bathe2006fem,belytschko2013nonlinear}. These factors may generate local instabilities and buckling modes that appear in an apparently random manner from one simulation to another.
            
    These instabilities may correspond to mechanically realistic buckling modes~\cite{timoshenko1961elastic}. Even when triggered by numerical discrepancies, dispersion acts as an initial perturbation that activates physically plausible instability mechanisms. Its numerical origin does not diminish the mechanical relevance of the observed deformation modes.
            
    Another major source of variability arises from computational optimizations related to hardware architectures, particularly CPUs. To reduce computation time, solvers exploit parallelization strategies and memory optimizations specific to each architecture. However, these optimizations alter the order of arithmetic operations, leading to rounding differences and cumulative numerical deviations~\cite{goldberg1991floating,collange2015numerical,higham2002accuracy}. Consequently, the same simulation executed on different architectures or with different optimization levels may produce slightly different results, which can then be amplified during the dynamic computation.
            
    The consequences of this dispersion are manifold. At the local scale, it can alter deformation modes and stress concentration regions. At the global scale, it may induce significant variations in performance indicators such as energy absorption, accelerations, or structural intrusions. Such discrepancies are likely to directly affect vehicle design and optimization decisions.
        
    \subsection*{}
        
    \begin{figure}[H]
        \centering
        \includegraphics[width=0.5\linewidth]{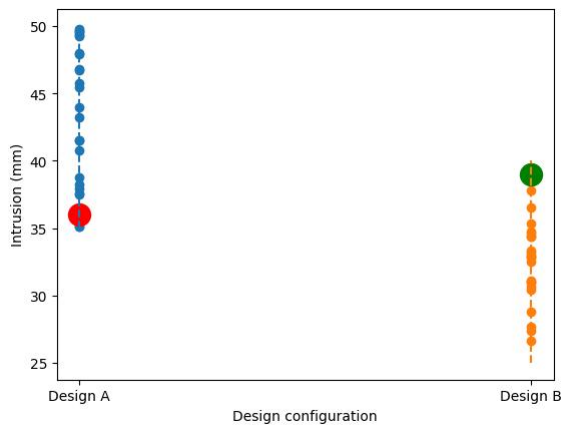}
        \caption{Illustration of the effect of numerical dispersion on design ranking based on a single simulation sample.}
        \label{fig:simple_exemple}
    \end{figure}
    
    \autoref{fig:simple_exemple} illustrates the risk associated with decisions based on a single simulation for each design configuration. Designs \textit{A} and \textit{B} are evaluated using an intrusion criterion (lower values are preferable). The dotted vertical lines represent the range of possible outcomes due to numerical dispersion, while the red and green dots correspond to single sampled simulation results.
    
    In this example, the sampled realization suggests that design \textit{A} outperforms design \textit{B}. However, considering the full dispersion range, the average intrusion associated with design \textit{B} is lower than that of design \textit{A}. A decision based solely on a single sample may therefore lead to an incorrect ranking of design alternatives.
    
    In an industrial context, such misclassifications may result in suboptimal design choices. Quantifying and accounting for numerical dispersion is therefore essential to improve the reliability of performance assessment and decision-making processes.
\section*{}

The remainder of this paper is organized as follows.
\hyperref[sec:tool]{Section~\ref*{sec:tool}} presents the industrial context and the objectives of the proposed tool.
\hyperref[sec:theory]{Section~\ref*{sec:theory}} introduces the theoretical framework underlying the methodology.
\hyperref[sec:results]{Section~\ref*{sec:results}} reports the experimental results and comparative analysis.
Finally, \hyperref[sec:conclusion_perspectives]{Section~\ref*{sec:conclusion_perspectives}} summarizes the main findings, discusses validation strategies, and outlines future perspectives.

\section{Reference Problem}
\label{sec:tool}

    This section introduces the crash scenario and the positioning of the proposed method with respect to robust design.
    
\begin{figure}[H]
        \centering
        \includegraphics[width=0.4\linewidth]{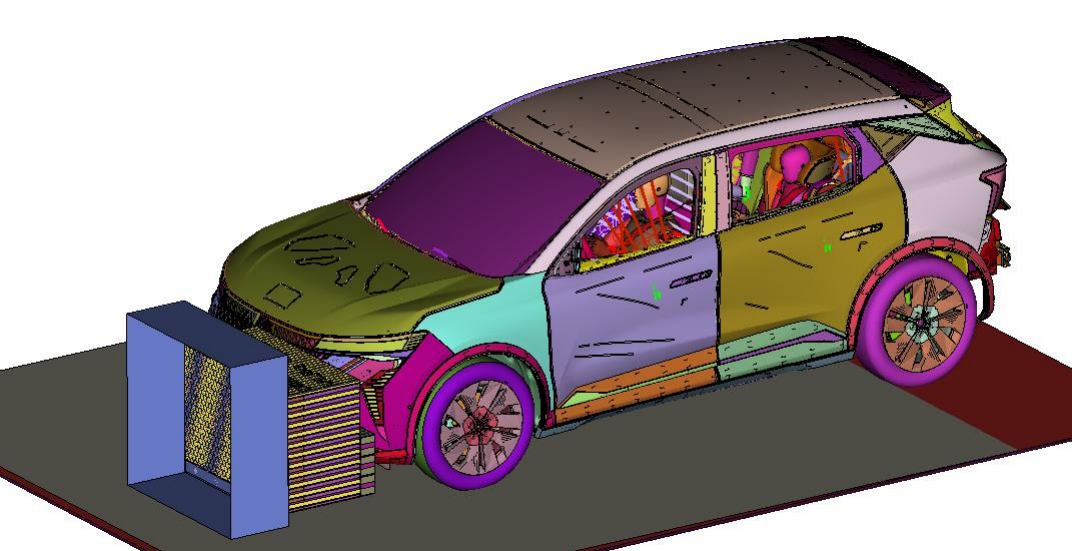}
        \caption{Finite element model of the vehicle at the initial time step (pre-impact configuration)}
        \label{fig:initial_model}
    \end{figure}

    The reference configuration considered in this study is illustrated in \autoref{fig:initial_model}, showing the vehicle model in its initial state prior to the crash simulation. The considered crash scenario corresponds to a frontal left impact at a velocity of 65\,km/h, faster than the standard regulatory crash conditions~\cite{euro_ncap2020,iihs2020}. 
    
    \subsection{Crash Scenario}

    The \CDC{} defines a target crash scenario corresponding to the expected structural behavior of the vehicle. In the case of a frontal impact, the nominal sequence of events can be described as follows~\cite{jones2010structural,belytschko2013nonlinear}:
    
    \begin{itemize}[label=-]
        \item Initial deformation of the front bumper beam and crash boxes;
        \item Compression of the front portion of the longitudinal member;
        \item Deployment of the front airbags;
        \item Controlled bending of the longitudinal member toward the nearest wheel in order to deflect the powertrain and preserve occupant compartment integrity (in particular, the footwell), without rotation of the A-pillar base;
        \item Transfer of loads from the wheel to the rocker panel;
        \item Maintenance of occupant compartment stability and compliance with regulatory criteria (e.g., HIC15 < 700, thoracic compression 42–60 mm, femoral force < 6 kN)~\cite{euro_ncap2020}.
    \end{itemize}
    
    The crash scenario cannot be reduced to the sole numerical values of performance criteria. The chronology of events and their sequencing are considered at least as important as the levels reached. In industrial practice, a slight increase in a given criterion may be acceptable if it occurs at the appropriate moment within the crash sequence~\cite{jones2010structural}.
    
    Each stage of the scenario is itself subject to numerical dispersion. This variability may affect not only the amplitude of structural responses, but also the timing of occurrence and the ordering of mechanisms such as local buckling, contact activation, or load transfer.
    
    The stability of the crash scenario therefore constitutes a requirement as critical as compliance with quantified performance criteria. A local deviation may alter load paths, disrupt energy redistribution, and lead to significantly different final responses. This justifies analyzing dispersion both at the level of individual \neu{}s and at the global scale of the crash scenario.

\subsection{Robust Design}
\label{sec:conception-robuste}

    Robust design aims to consider performance criteria as statistical distributions rather than single nominal values~\cite{taguchi1986introduction}. Its objective is to ensure compliance with requirements despite manufacturing and material uncertainties.
    
    In the automotive industry, this philosophy is already partially embedded in the definition of the \CDC{}. The selected crash scenarios generally correspond to severe cases, conservatively defined to maximize the robustness of structural behavior with respect to industrial variability~\cite{jones2010structural}.
    
    Manufacturing tolerances and variations in material properties introduce dispersion in performance indicators. Design parameters primarily influence the mean value of the response, whereas tolerances affect its variance. Performance dispersion therefore intrinsically depends on design choices and varies across the solution space~\cite{taguchi1986introduction}.
    
    Robust design methods cover a broad spectrum, ranging from Taguchi’s statistical approaches to reliability-based optimization methods such as FORM and SORM~\cite{rackwitz1978structural,breitung1984asymptotic}.
    
    The present work addresses a distinct source of uncertainty: numerical dispersion arising from crash computation. This dispersion appears independently of geometric or material variations, even when initial conditions are strictly identical.
    
    The proposed method does not replace classical robust design approaches; rather, it complements them by enabling the identification and quantification of the crash model’s sensitivity to numerical dispersion, thereby enhancing the reliability of result interpretation and decision-making.

\subsection{Tool Objectives}

    The objective is to identify structural regions sensitive to numerical dispersion at the mesh level, particularly in areas associated with requirement criteria.
    
    The tool operates as a post-processing diagnostic, detecting locations where small numerical perturbations may significantly alter structural response and affect design decisions. For example, dispersion in footwell intrusion may lead to unnecessary reinforcement and mass increase.
    
    The methodology relies on a limited number of simulations and must balance detection capability with computational cost. It is designed to operate on everyday simulations already performed by designers, without requiring additional runs.
    
    Model lifetime is a critical issue. Since the \DT{} evolves continuously during development, the tool must remain effective under moderate design changes without frequent retraining. This raises the question of the minimum number of simulations required to ensure robustness and long-term validity.

    \begin{remark}
    The term \DT{} here denotes the complete numerical configuration of the model (geometry, mesh, material laws, contact definitions, and boundary conditions). In other companies, it may be commonly referred to as \textit{model configuration}, \textit{simulation setup}, or \textit{numerical model definition}.
    \end{remark}
\section{Theoretical Framework of the Proposed Solution}
\label{sec:theory}

    \subsection{Data and Study Framework}
    \label{dispersion_def}
    
        The data used in this study are obtained from crash simulations performed multiple times under strictly identical initial conditions. Despite this, differences appear in the results, particularly in the trajectories of certain model nodes. These nodes are classified into two categories: dispersed (exhibiting significant variation across runs) and non-dispersed (showing stable trajectories). Such variability under identical conditions is characteristic of complex nonlinear numerical systems~\cite{strogatz2018nonlinear}.
        
        All kinematic quantities are expressed in a vehicle-fixed Cartesian coordinate system, where:
    
        \begin{figure}[ht]
            \centering
            \begin{minipage}[t]{0.55\linewidth}
            \vspace{0pt}
            \begin{itemize}[label=-]
                \item the $x$-axis denotes the longitudinal direction of motion
                \item the $y$-axis denotes the lateral direction (positive to the left)
                \item the $z$-axis denotes the vertical direction (positive upward)
            \end{itemize}
            \end{minipage}
            \hfill
            \begin{minipage}[t]{0.4\linewidth}
            \vspace{0pt}
            \centering
            \includegraphics[width=0.5\linewidth]{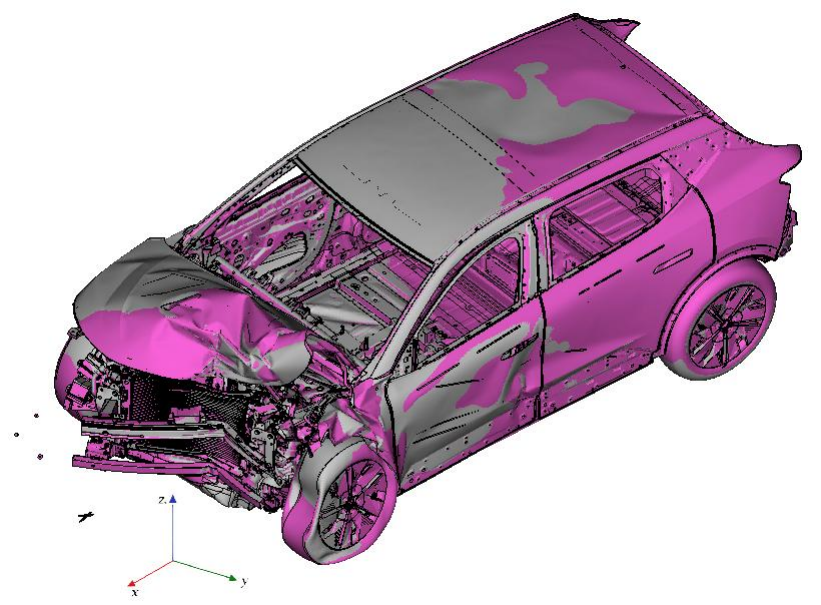}
            \caption{Vehicle-fixed coordinate system at peak crash time.}
            \label{fig:coordinate_system}
            \end{minipage}
        \end{figure}
    
        Dispersion is defined as spatial variability exceeding 5\,mm along the $x$-axis at peak crash response across repeated simulations. This threshold was selected based on engineering considerations: the order of magnitude of footwell intrusion in the considered crash configuration is approximately 50\,mm~\cite{jones2010structural}. A variability threshold corresponding to 10\% of this magnitude was therefore adopted. Although partially arbitrary, this value reflects a physically meaningful scale relative to the expected structural response.

        In a theoretical scenario where budgetary and computational constraints are absent, it would be possible to repeat each crash simulation a large number of times in order to exhaustively identify regions sensitive to numerical dispersion and to integrate this analysis into a robust design framework~\cite{taguchi1986introduction}. In such a case, the proposed tool would be unnecessary, as it would suffice to measure the distribution of observables, such as displacement or velocity, for each mesh node.
        
        In industrial contexts, however, this approach is impractical because of its computational cost, as it would require multiplying the simulation budget by a factor of 10 or 20. This limitation motivates the development of methods capable of detecting sensitive regions from a restricted number of simulation runs.
    
    \subsection{Approach 1: Supervised Learning from Raw Input Data}
    
        This approach relies on a supervised learning model designed to classify \neu{}s into two categories: dispersed and non-dispersed~\cite{hastie2009elements}.
        
        The dataset consists of 15 repetitions of the same simulation. Ten simulations are used for training and five for validation.
        
        The inputs correspond to the complete temporal trajectories of the \neu{}s. The trajectories comprise 29 time steps, corresponding to the industrial standard for result storage in crash simulations. This discretization reflects common post-processing practices in an operational context and corresponds to a coarse sampling of the underlying dynamic signal~\cite{oppenheim1999signals}.
        
        Trajectories are expressed in terms of displacements relative to the initial configuration. Absolute coordinates are directly affected by geometric modifications introduced during vehicle development or by architectural differences between distinct vehicles. Displacements, expressed relative to the initial state, exhibit reduced dependence on the \DT{}.
        
        Each trajectory is considered as a dynamic signature of the \neu{} behavior during the crash. The selected classifier is a Random Forest~\cite{breiman2001random}. This method aggregates multiple decision trees trained on random subsamples~\cite{breiman2001random,biau2015random}. It mitigates overfitting and demonstrates good robustness to numerical noise. It is well suited to high-dimensional temporal signals.

        The Random Forest is used as a baseline model due to its ease of implementation and strong performance in supervised classification tasks~\cite{hastie2009elements}. This choice will be discussed later in comparison with approaches that are less dependent on the \DT{}.
        
        Dependence on the \DT{} nevertheless remains significant. Any modification of the geometry, mesh, or material parameters alters the distribution of input data and may render the model obsolete, requiring retraining. This constraint limits the long-term integration of the approach in an industrial context where the \DT{} evolves daily.
    
    \subsection{Approach 2: Supervised Learning Using Data Independent From The \textit{Technical Definition}}
    \label{DT_inde}
    
        It was necessary to increase the sampling frequency of the signal to improve the temporal resolution of the analysis. To this end, the number of time steps saved during the simulation was increased from 29 to 289. This provides a more detailed description of trajectory evolution and facilitates the detection of subtle signal variations that might be masked at lower temporal resolution~\cite{oppenheim1999signals}.
        
        A limitation of this approach is that the sampling frequency could not be further increased due to storage constraints (167~GB per run for 289 time steps; 10 runs correspond to 1.67~TB). This trade-off between temporal accuracy and data volume constitutes a practical limitation of the method and must be taken into account when interpreting the results. These constraints also led to a reduction in the number of simulations performed, from 15 to 10.
    
    \subsubsection{Mathematical Decomposition of Trajectories Using Fourier and Wavelet Transforms as Explanatory Variables of Dispersion}
    
        Trajectory signals are projected into the frequency domain using Fourier and wavelet transforms~\cite{abry2019,oppenheim1999signals}. These decompositions are used to extract frequency-based indicators of numerical dispersion.
        
        The Fourier transform provides a global frequency representation of the signal, enabling the identification of persistent spectral components. However, it does not capture the temporal localization of these components.
        
        Wavelet transforms provide a joint time–frequency representation, allowing the detection of transient events and localized instabilities. This makes them particularly suitable for identifying dispersion patterns that emerge during specific phases of the crash event~\cite{mallat1999wavelet}.
        
        In this study, the analysis is performed using a Discrete Wavelet Transform (DWT) with a multi-level decomposition implemented through the \texttt{wavedec} function from the PyWavelets library. The selected mother wavelet belongs to the Daubechies family, which provides a good compromise between compact support and time–frequency localization~\cite{mallat1999wavelet}. More specifically, the db4 wavelet is used. It is a commonly used wavelet in practice, offering a good trade-off between sensitivity to local variations and robustness to noise.
        
        Unlike the Fourier transform, which provides a global frequency representation of the signal, wavelet analysis makes it possible to capture transient variations localized in time. This property is particularly relevant for crash simulations, where numerical dispersion typically appears during short dynamic phases associated with contact events or structural instabilities.
        
        The working hypothesis is that recurrent spectral components, particularly low-amplitude modes observed only in dispersed nodes, constitute characteristic dispersion signatures.
        
        Because these representations rely on signal dynamics rather than geometric descriptors, they are less sensitive to variations in the \DT{}. This facilitates the construction of transferable dispersion indicators across model configurations.

    \subsubsection{Temporal Slope Variation as an Explanatory Variable of Dispersion}
    
\begin{figure}[H]
        \centering
        \includegraphics[width=0.7\linewidth]{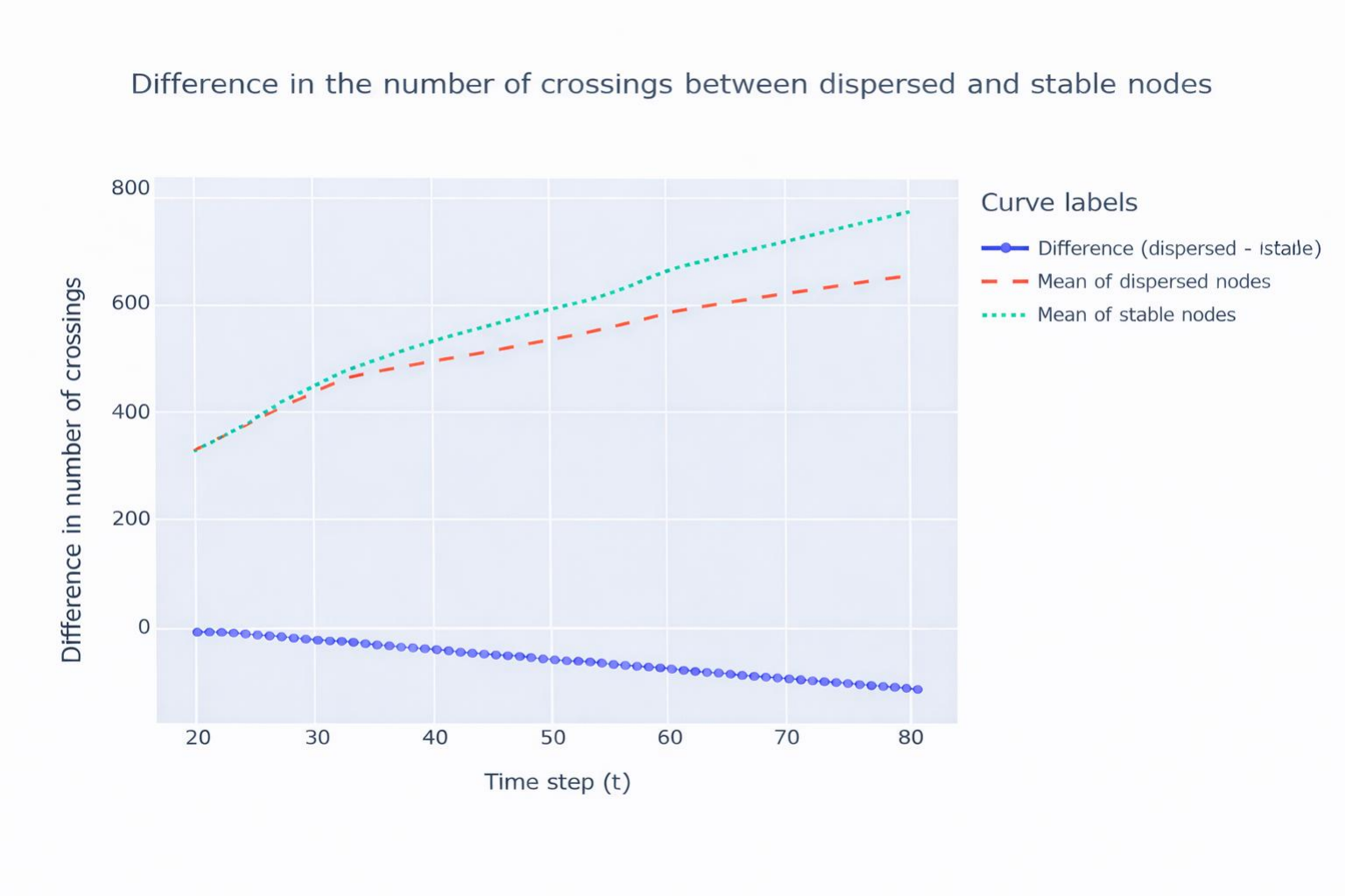}
        \caption{Evolution of the number of intersections over time for a dispersed node and a non-dispersed node}
        \label{fig:croisement_de_noeuds}
        \end{figure}

        An alternative indicator is derived from the cumulative number of intersections between the trajectories of the same node across multiple simulation runs. Dispersed and non-dispersed nodes exhibit distinct intersection patterns (Figure~\ref{fig:croisement_de_noeuds}).
        
        Non-dispersed nodes show strongly overlapping trajectories, resulting in a high number of intersections. Dispersed nodes progressively diverge, leading to a reduction in intersections as numerical discrepancies amplify over time~\cite{strogatz2018nonlinear}.
        
        This motivates a local analysis based on slope variations between successive time steps. By focusing on local trajectory evolution, dispersion onset can be detected while reducing sensitivity to the \DT{}. The resulting indicator is formulated independently of mesh topology and geometric parameterization.

        In crash simulations, the actual time integration is performed over several hundred thousand time steps. As a result, referring to temporal derivatives may be misleading at the available sampling resolution, and describing the signal evolution in terms of local slope variations provides a more appropriate and consistent interpretation.
    \subsection{Node Classification with a Rank Reduction Autoencoder (RRAE)}

Neural networks are used to combine nonlinear feature extraction with supervised classification in order to identify dispersed nodes from dynamic signals~\cite{goodfellow2016deep}. The Random Forest model introduced previously serves as a baseline. However, it operates directly on raw inputs, depends on the statistical distribution of the training data, and does not structure the underlying information.
    
The proposed framework relies on a Rank Reduction Autoencoder (RRAE) coupled with a Multi-Layer Perceptron (MLP). The RRAE, introduced in~\cite{rrae2024}, provides a nonlinear model-order reduction framework that combines autoencoder architectures with explicit low-rank constraints in the latent space. This formulation enables the extraction of structured latent representations while preserving the nonlinear relationships inherent in complex signals.
    
RRAE has recently been extended to a variety of engineering applications. In particular, it has been applied to generative design and inverse modeling tasks~\cite{rrae_generative_design}, as well as to real-time parametric geometry generation~\cite{rrae_gpd}. It has also demonstrated strong capabilities in mechanical design optimization and nonlinear dynamical system modeling~\cite{rrae_topology,rrae_rraedy}. Importantly for the present work, RRAE has been successfully used for damage detection and structural health monitoring, where it enables the identification of subtle patterns associated with structural anomalies~\cite{rrae_damage}.
    
The RRAE is therefore considered in this work as a promising representation-learning approach. Its ability to capture dominant patterns of variation enables the extraction of structured latent features that better represent intrinsic signal variations. Such structured representations have also been exploited for feature discovery and physical characterization in complex engineering datasets~\cite{rrae_surface}. Therefore, the objective is to learn a structured latent representation of node dynamics prior to classification, following the general principles of representation learning~\cite{bengio2013representation}.
    
An autoencoder (AE) is a neural network trained in a self-supervised manner to learn a compact latent representation of an input signal by minimizing the reconstruction error between the input and its reconstruction~\cite{hinton2006reducing,goodfellow2016deep}. It consists of:
    
\begin{itemize}
    \item an encoder that maps the input data to a reduced-dimensional latent space,
    \item a decoder that reconstructs the original signal from this latent representation.
\end{itemize}

The training procedure takes as input a dataset $X_{\text{train}}$ composed of node response signals, together with the corresponding target labels $Y_{\text{train}}$ used for supervised learning. Each target label is binary: $y \in \{0,1\}$, where $y=1$ denotes a dispersed node and $y=0$ denotes a non-dispersed node.

At the sample level, each input is represented as a vector $x = (x_1, x_2, \dots, x_D) \in \mathbb{R}^D$, where each component corresponds to a feature describing the node response. Depending on the chosen representation, these components may correspond to sampled values of a physical quantity over time (e.g., displacement), spatial coordinates, or derived descriptors computed from the signal. In the case of trajectory-based inputs, $x$ represents the discretized signal over $D$ time steps.
    
The encoder maps this input to a latent representation denoted by $z \in \mathbb{R}^L$. The Rank Reduction Autoencoder (RRAE)~\cite{rrae2024} introduces an explicit rank constraint in the latent space, enforced through a truncated singular value decomposition (SVD) or an equivalent penalization term in the loss function. This low-rank constraint enables the separation of dominant coherent structures from noise, a property that is central to the effectiveness of RRAE in nonlinear system modeling~\cite{rrae2024,rrae_rraedy}.
    
The encoded representation $z$ is projected onto a low-dimensional subspace using truncated SVD, yielding a reduced latent representation $z_r \in \mathbb{R}^{k_{\max}}$, which corresponds to the projection onto the dominant singular modes~\cite{golub2013matrix}. This constraint is closely related to classical dimensionality reduction techniques such as principal component analysis (PCA), while extending them to nonlinear representations~\cite{jolliffe2016pca}. Compared with purely linear approaches, RRAE enables the extraction of structured low-rank representations while preserving nonlinear relationships in the data, which has motivated its use in applications such as damage detection and structural health monitoring~\cite{rrae_damage}. It favors structured patterns of variation while filtering out noisy or unstable components, improving robustness to numerical perturbations and moderate variations of the \DT{}.
    
The decoder reconstructs the input from the reduced latent representation, yielding $\hat{x} = (\hat{x}_1, \hat{x}_2, \dots, \hat{x}_D)$, while the classifier predicts an output $\hat{y}$ from the same reduced representation.
    
The main hyperparameters of the model are the retained rank $k_{\max}$, which controls the dimensionality of the reduced latent space, the weighting coefficients $\lambda_{\text{recon}}$ and $\lambda_{\text{cls}}$, which balance the reconstruction and classification objectives in the joint loss function, and the numbers of training epochs $N_1$ and $N_2$, corresponding respectively to the joint training phase and the fine-tuning phase. These parameters directly influence the trade-off between information preservation, noise filtering, and the discriminative capability of the learned representation.

\begin{figure}[h!]
\centering
\includegraphics[width=0.8\linewidth]{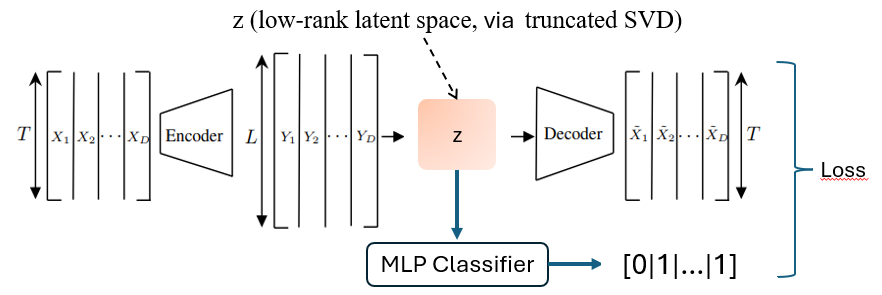}
\caption{Schematic representation of the RRAE--MLP architecture.}
\label{fig:AE_MLP}
\end{figure}

\begin{figure}[h!]
\centering

\begin{tcolorbox}[colback=white,colframe=black,title=Rank Reduction Autoencoder (RRAE) with MLP]

\begin{algorithmic}[1]

\State \textbf{Input:} $X_{\text{train}}, Y_{\text{train}}$
\State \textbf{Hyperparameters:} $k_{\max}, \lambda_{\text{recon}}, \lambda_{\text{cls}}, N_1, N_2$

\State Initialize Encoder, Decoder, and MLP
\State Initialize optimizer

\Statex \textbf{Phase 1 --- Joint training}
\For{epoch $= 1$ to $N_1$}
    \For{each mini-batch $(x, y)$}
        \State $z \gets \mathrm{Encoder}(x)$
        \State $z_r \gets \mathrm{SVDProj}(z, k_{\max})$
        \State $\hat{x} \gets \mathrm{Decoder}(z_r)$
        \State $\hat{y} \gets \mathrm{MLP}(z_r)$

        \State $L_{\text{recon}} \gets \mathrm{reconstruction\_loss}(\hat{x}, x)$
        \State $L_{\text{cls}} \gets \mathrm{classification\_loss}(\hat{y}, y)$
        \State $L_{\text{total}} \gets \lambda_{\text{recon}} L_{\text{recon}} + \lambda_{\text{cls}} L_{\text{cls}}$

        \State Backpropagate($L_{\text{total}}$)
        \State Update Encoder, Decoder, and MLP parameters
    \EndFor
\EndFor

\Statex \textbf{Phase 2 --- Compute the fixed low-rank basis}
\State Compute latent representations $Z = \mathrm{Encoder}(X_{\text{train}})$
\State Compute truncated SVD of $Z$
\State Keep the first $k_{\max}$ singular vectors to form the basis $U_{k_{\max}}$
\State Freeze $U_{k_{\max}}$

\Statex \textbf{Phase 3 --- Fine-tuning with fixed basis}
\For{epoch $= 1$ to $N_2$}
    \For{each mini-batch $(x, y)$}
        \State $z \gets \mathrm{Encoder}(x)$
        \State $z_r \gets \mathrm{Project\_onto\_fixed\_basis}(z, U_{k_{\max}})$
        \State $\hat{x} \gets \mathrm{Decoder}(z_r)$
        \State $\hat{y} \gets \mathrm{MLP}(z_r)$

        \State $L_{\text{recon}} \gets \mathrm{reconstruction\_loss}(\hat{x}, x)$
        \State $L_{\text{cls}} \gets \mathrm{classification\_loss}(\hat{y}, y)$
        \State $L_{\text{total}} \gets \lambda_{\text{recon}} L_{\text{recon}} + \lambda_{\text{cls}} L_{\text{cls}}$

        \State Backpropagate($L_{\text{total}}$)
        \State Update Decoder and MLP parameters while keeping Encoder and $U_{k_{\max}}$ fixed
    \EndFor
\EndFor

\end{algorithmic}

\end{tcolorbox}

\caption{Pseudocode of the RRAE--MLP training procedure.}
\label{fig:rrae_algo}
\end{figure}

As illustrated in Figure~\ref{fig:AE_MLP}, the encoded representation $z$ is projected onto a reduced latent space $z_r$, which is used both for reconstruction of the input $\hat{x}$ and for classification through a Multi-Layer Perceptron (MLP).
    
The MLP is a fully connected neural network with nonlinear activation functions, suited for classification in structured representation spaces~\cite{rumelhart1986learning}. It operates on the reduced latent representation, which encodes the dominant structured features of the input signal.
    
Training is performed in two stages. In the first stage, the encoder, decoder, and MLP are trained jointly by minimizing reconstruction and classification losses simultaneously. In the second stage, a fixed low-rank basis is computed from the latent representations, and the model is fine-tuned while keeping this basis fixed. In this way, the latent representation evolves to improve both signal reconstruction and class separability.
    
The RRAE structures the intrinsic variations of node responses, while the MLP exploits this representation to separate dispersed and non-dispersed nodes.
    
This framework can be trained from data generated on a single technical-definition configuration, using repeated simulations of the same crash case, which is compatible with industrial constraints. Because classification is performed on structured latent features rather than raw inputs, sensitivity to moderate variations of the technical definition is expected to be reduced. This may limit the need for retraining under incremental geometric or material modifications. The proposed RRAE-based framework combined with an MLP classifier therefore provides a promising approach for improving robustness, generalization capability, and stability in dispersion detection for crash post-processing, in line with recent advances demonstrating its effectiveness in engineering design and predictive modeling contexts~\cite{rrae_generative_design,rrae_topology}.
\section{Results}
\label{sec:results}

\subsection{Data Preprocessing}

    The preprocessing procedure is identical for all approaches and consists of progressively restricting the analysis to regions that are relevant to the crash response.
    
    Components not directly involved in impact dynamics, such as the front hood sheet metal and certain door elements, are removed to reduce noise. Rear components are excluded, as the study focuses on a front-left crash configuration.
    
\begin{figure}[ht]
        \centering
        \includegraphics[width=0.3\linewidth]{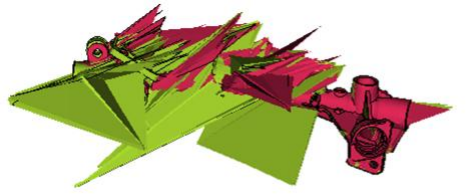}
        \caption{Superposition of the steering rack housing obtained from two identical simulation runs, shown in red and green, highlighting significant numerical dispersion. The triangular points correspond to calculation errors, which lead to some nodes being left behind while still being displayed. The two runs produce different calculation errors even though they have the same initial conditions.}
        \label{fig:corps_de_cremaillere}
    \end{figure}

    Some parts exhibiting abnormal behavior between identical runs are also removed. For instance, the steering rack housing shows significant variability (\autoref{fig:corps_de_cremaillere}), which could bias the analysis.
    
    Rigid body motions were removed to retain only structural deformations, thereby isolating energy absorption and numerical dispersion effects from global vehicle motion.
    
    The dataset is balanced by keeping equal numbers of dispersed and non-dispersed nodes.
    
    Dispersion is defined as in \autoref{dispersion_def}. It is quantified by comparing repeated simulations for each node. An initial dataset of 15 runs is used for baseline analyses in order to better characterize dispersion variability. For subsequent analyses, the number of runs is reduced to 10, which provides a compromise between statistical representativeness and practical constraints, including computational cost, data storage limitations (\autoref{DT_inde}), and the need for balanced training and validation splits.
    
\subsection{Supervised Learning Using Raw Data}

    After applying the data preparation procedure, a supervised learning approach was implemented to evaluate the model’s ability to classify dispersed nodes. The data were divided into two sets: 10 runs for training and 5 runs for validation.
    
    The model inputs correspond to the temporal trajectories of the nodes, i.e., the evolution of their positions over time during the crash simulation. From these time series, the model aims to predict a binary label, taking the value 1 for a dispersed node and 0 for a non-dispersed node (defined using the full training dataset).
    
    Cross-validation was employed in order to strengthen the robustness of the evaluation~\cite{hastie2009elements}.
    
\begin{figure}[H]
        \centering
        \includegraphics[width=0.4\linewidth]{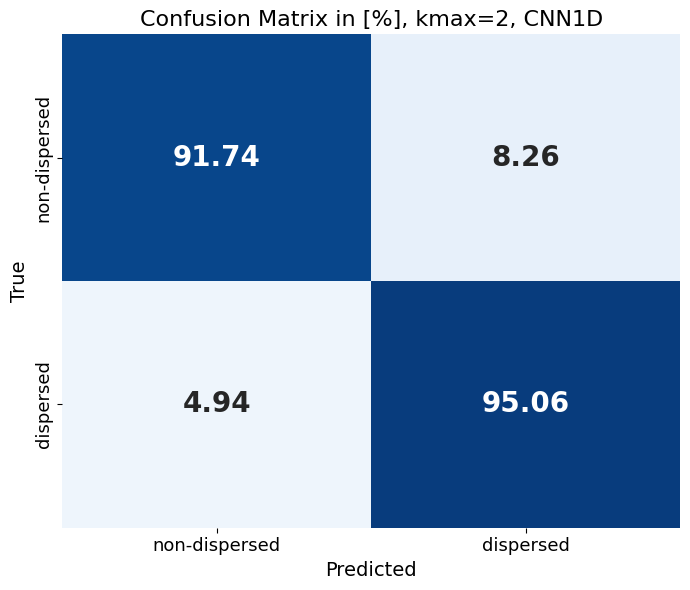}
        \caption{Confusion matrix obtained with the supervised learning approach with raw data}
        \label{fig:confusion_classic}
    \end{figure}

    The results show an accuracy above 90\% (\autoref{fig:confusion_classic}), confirming that a standard supervised model is already able to identify a large proportion of dispersed nodes from raw trajectories. More precisely, 91.74\% of non-dispersed nodes and 95.06\% of dispersed nodes are correctly classified. This performance validates the use of Random Forest as a baseline approach while also showing that there is still room for improvement, particularly in reducing confusion between the two classes and improving robustness with respect to the \DT{}.
    
\subsection{Classification Based on Data Independent of the \textit{Technical Definition}}

    \subsubsection{Fourier and Wavelet Decomposition}
    
    In this approach, an analytical framework is adopted in order to reduce dependence on the \DT{} of the model. The principle is based on the mathematical decomposition of the temporal trajectories of the nodes using Fourier and wavelet transforms, whose concepts were detailed in the previous section. These methods analyze the frequency content of the signal and identify characteristic signatures associated with numerical dispersion~\cite{oppenheim1999signals,mallat1999wavelet}.
    
    In this study, the transforms are applied both to the nodal displacement signals and to their local slope variations between successive time steps. This allows the analysis to capture both the global evolution of the trajectories and the local dynamic variations associated with dispersion phenomena.
    
    Fourier and wavelet transforms are purely signal-based operators applied to temporal trajectories. They do not depend on geometric coordinates, mesh topology, or part identifiers. The extracted descriptors characterize dynamic behavior through spectral content and time--frequency localization, independently of the spatial discretization of the model.
    
    Moderate modifications of geometry, mesh density, or architectural layout may alter absolute trajectory amplitudes, but they do not fundamentally modify the spectral structure of the response as long as the underlying crash mechanisms remain comparable~\cite{oppenheim1999signals}. For this reason, frequency-based indicators derived from these transforms exhibit reduced sensitivity to variations in the \DT{} and improved transferability across configurations.
    
    For the following analyses, the total number of runs was reduced from 15 to 10 in order to adopt a balanced training and validation strategy, with a 50/50 split between the training set and the validation set. In addition, to improve temporal resolution, the number of time steps was increased from 29 to 289 sampling points.
    
    \subsubsection*{}
    \begin{figure}[H]
        \centering
        \includegraphics[width=0.4\linewidth]{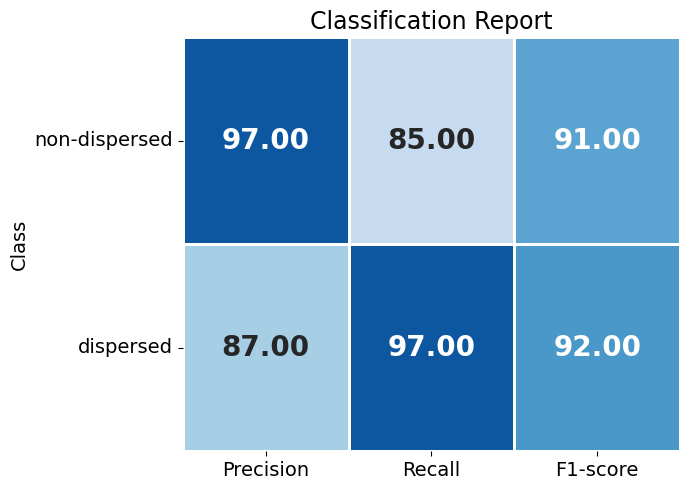}
        \caption{Classification performance obtained from features extracted using the Fourier transform.}
        \label{fig:confusion_fft}
    \end{figure}
    
    The results in \autoref{fig:confusion_fft} show that Fourier-based features already provide meaningful discriminative information. The recall reaches 97\% for dispersed nodes, indicating that the model is able to detect most unstable responses and produces relatively few false negatives. However, the recall drops to 85\% for non-dispersed nodes, which reveals a less balanced classification. This suggests that global spectral features capture dispersion-related signatures, but remain insufficient to fully separate stable and unstable behaviors when their frequency content partially overlaps.
    
    \subsubsection*{}
    \begin{figure}[H]
        \centering
        \includegraphics[width=0.4\linewidth]{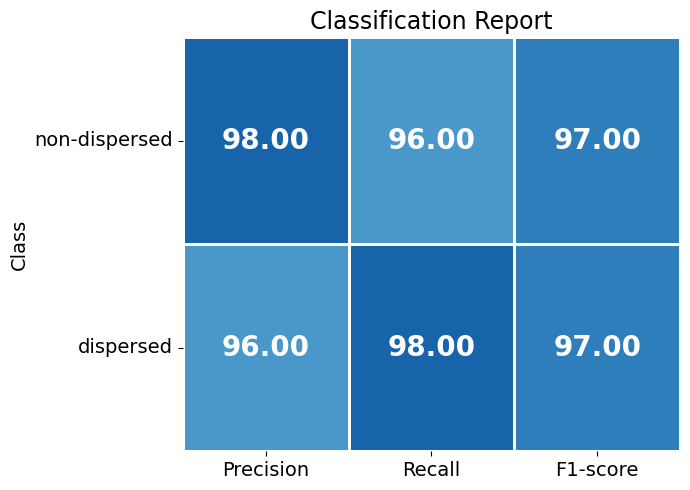}
        \caption{Classification performance obtained from features extracted using the wavelet transform.}
        \label{fig:confusion_ondelette}
    \end{figure}
    
    The results in \autoref{fig:confusion_ondelette} show a clear improvement when wavelet-based features are used. Precision and recall reach 96--98\% for both classes, yielding a much more balanced classification than with Fourier features. This indicates that the wavelet representation captures not only the spectral structure of the signal but also its temporal localization, which is essential for identifying transient events associated with contact changes, local instabilities, and dispersion onset during the crash sequence.
    
    \subsubsection*{}
    
    In comparison, the wavelet-based approach appears more robust and more discriminative than the Fourier transform. This result highlights the importance of combining temporal and frequency information when analyzing crash signals. Numerical dispersion is often associated with localized dynamic events, which cannot be fully described by a purely global spectral representation. The superior performance of wavelet features therefore confirms the relevance of time--frequency analysis for detecting numerical dispersion.
    
    \subsubsection{Temporal Slope Variation}
    
    The analysis is restricted to three components: two footwell components and the \textit{front cross-member}. Node trajectories are transformed into local slope variations between successive time steps and truncated at 220 time steps, corresponding to the compression phase.
    
    \begin{figure}[H]
        \centering
        \includegraphics[width=0.4\linewidth]{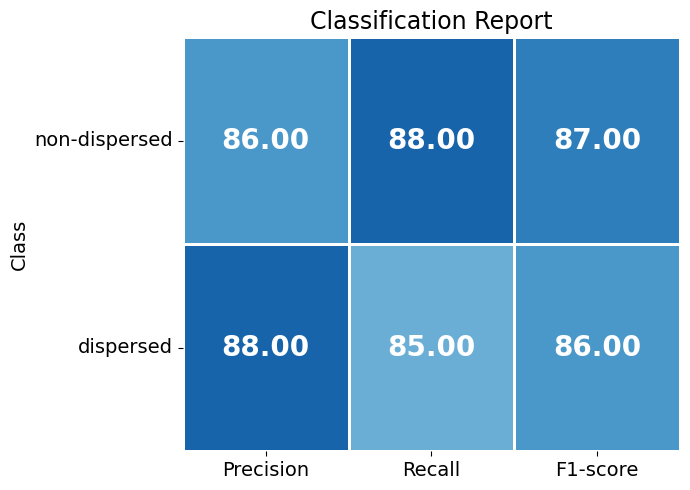}
        \caption{Classification performance obtained from temporal slope variations.}
        \label{fig:pente_results}
    \end{figure}
    
    \autoref{fig:pente_results} reports the classification performance obtained from slope-based descriptors. The precision for the dispersed class is 0.87, while the overall scores remain around 86--88\% for both classes. These results confirm that local slope variations contain useful information on the onset of instability, since they directly emphasize abrupt temporal changes in the signal. However, the performance remains below that obtained with wavelet-based features, suggesting that local differential information alone does not provide a sufficiently rich description of the signal and may miss part of the global dynamic context.

\subsection{RRAE-Based Classification with MLP}

    This section presents a classification framework based on a Rank Reduction Autoencoder (RRAE) combined with a Multi-Layer Perceptron (MLP) for the identification of dispersed nodes using different signal representations. Unlike previous approaches, training is performed using a single simulation, which constitutes a strong constraint in an industrial context. The objective is therefore to evaluate the ability of the learned latent representation to generalize from limited data while preserving discriminative information.
    
    \subsubsection{Position-based Input}
    
    The temporal trajectories of the nodes are projected into the latent space learned by the RRAE and then classified by the MLP.
    
    \begin{figure}[H]
    \centering
    \includegraphics[width=0.4\linewidth]{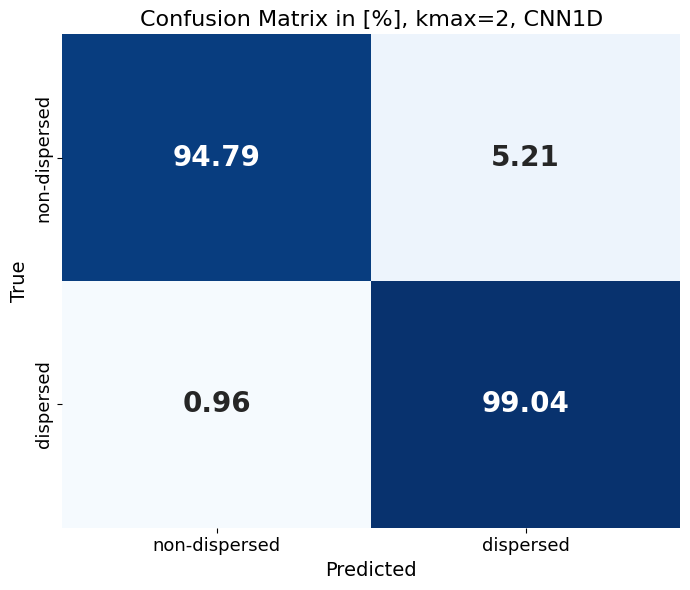}
    \caption{Confusion matrix obtained using position trajectories.}
    \label{fig:confusion_rrae_pos}
    \end{figure}
    
    The results in \autoref{fig:confusion_rrae_pos} show that the RRAE-based framework already improves classification when using raw position trajectories as input. The recognition rates reach 94.79\% for non-dispersed nodes and 99.04\% for dispersed nodes, which is higher than the Random Forest baseline on raw coordinates. This indicates that the latent representation learned by the RRAE captures the dominant structures of the signal while preserving class-relevant information. In particular, the very high recognition rate for dispersed nodes suggests that the model effectively identifies instability-related patterns. The slightly lower performance for non-dispersed nodes indicates that some borderline cases remain difficult to separate, but the overall result confirms the benefit of structured latent learning compared with direct classification on raw inputs.
    
    \subsubsection{Wavelet-Based Input (DWT)}
    
    The trajectories are transformed using a Discrete Wavelet Transform (DWT) in order to capture localized time--frequency features while reducing dependence on the \DT{}. In this study, a Daubechies wavelet of order 4 (db4) with multi-level decomposition is used~\cite{mallat1999wavelet}.
    
    Due to the comparatively weaker performance of Fourier-based features in the Random Forest classification, only DWT-based inputs are retained in the RRAE-based classification.
    
\begin{figure}[H]
    \centering
    \includegraphics[width=0.4\linewidth]{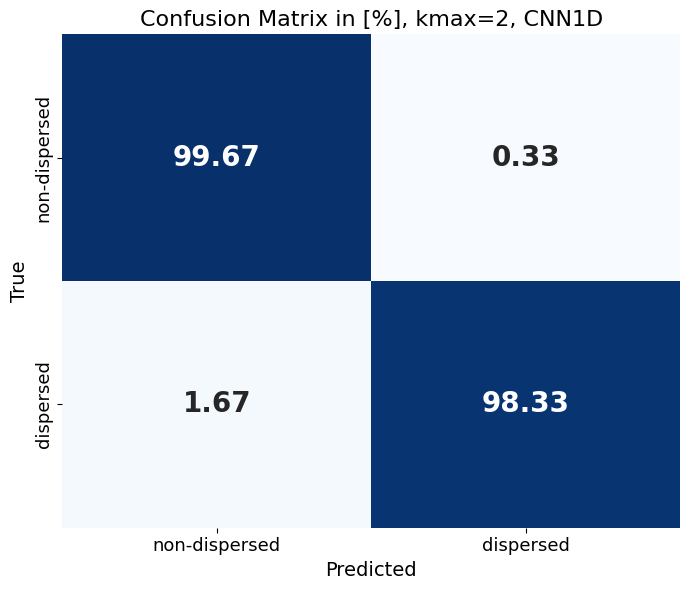}
    \caption{Confusion matrix obtained using wavelet-based input (DWT, db4).}
    \label{fig:confusion_rrae_wavelet}
    \end{figure}

    This representation emphasizes transient dynamics and instability patterns associated with numerical dispersion, while remaining less sensitive to global geometric variations. \autoref{fig:confusion_rrae_wavelet} presents the results.
    
    The results show excellent classification performance, with 99.67\% of non-dispersed nodes correctly classified and 98.33\% of dispersed nodes correctly identified. This confirms that the latent representation learned from wavelet-based inputs preserves the discriminative information required for node classification. The very high recognition rates in both classes indicate that the RRAE is able to organize transient signal characteristics into a compact representation that remains highly informative for the MLP. Compared with raw position-based inputs, the wavelet representation appears better suited to isolating short-duration events associated with local instabilities, contact changes, or rapid structural response variations. These results therefore support the relevance of combining time--frequency preprocessing with rank-reduced latent learning.
    
    \subsubsection{Slope-Variation-Based Input}
    
    Slope variations between successive time steps are used as input. Trajectory crossings are not practically exploitable, as they would require repeating the simulation multiple times each time the tool is applied. This representation highlights rapid dynamic changes in the signal and therefore focuses directly on the local temporal irregularities associated with dispersion.
    
    \begin{figure}[H]
    \centering
    \includegraphics[width=0.4\linewidth]{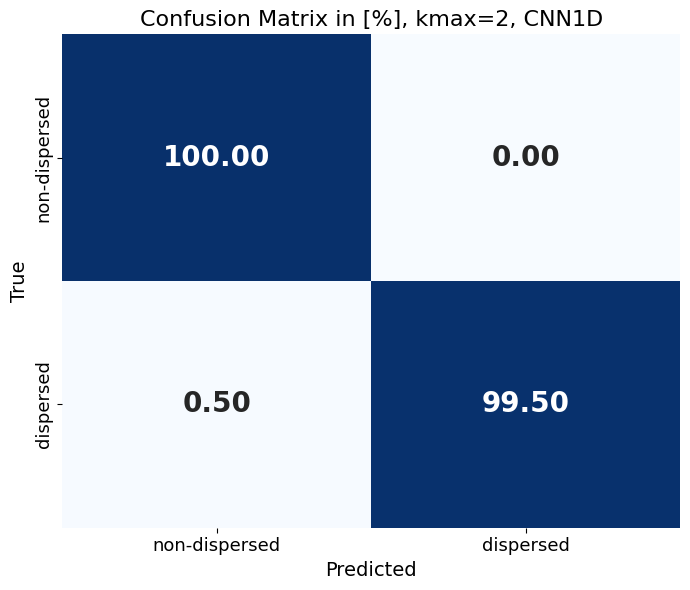}
    \caption{Confusion matrix obtained using slope variations.}
    \label{fig:confusion_rrae_slope}
    \end{figure}
    
    The results shown in \autoref{fig:confusion_rrae_slope} correspond to the best performance obtained among the tested input representations. Classification remains highly accurate for the non-dispersed class, while the recognition rate for dispersed nodes reaches 99.50\%. This suggests that slope variations provide a particularly effective description of the local signal changes associated with the onset and amplification of numerical dispersion. Compared with raw trajectories, which also contain global motion trends, slope-based inputs emphasize the rapid local variations that are most relevant for instability detection. Combined with the low-rank latent representation learned by the RRAE, this encoding leads to an almost complete separation between stable and unstable responses.
    
    \subsubsection{Summary}
    
    Overall, the results show that transformed signal representations improve classification performance compared with raw position-based inputs. This trend is already visible for the Random Forest baseline, for which wavelet-based features clearly outperform Fourier features and slope-based descriptors. However, the gains become much more significant when these representations are coupled with the RRAE-based framework.
    
    Among all tested encodings, the most effective are those that reduce dependence on the \DT{} while preserving the dynamic signatures associated with dispersion. Wavelet-based inputs provide the best compromise between local temporal sensitivity and global frequency characterization, leading to excellent and balanced performance for both classes. Slope-variation-based inputs yield the highest recognition rate for dispersed nodes and the best overall classification performance, suggesting that local differential information is especially relevant for identifying numerical instability patterns when combined with rank-reduced latent learning.
    
    \begin{table}[ht]
        \centering
        \label{tab:comparaison_codages}
        \begin{tabular}{l p{2cm} p{2.5cm}}
        \hline
        \textbf{Encoding} & \textbf{Multiple simulations required} & \textbf{Dependence on \DT{}} \\
        \hline
        Coordinates in x,y,z & \textbf{\textit{Yes}} & Very High \\
        Displacement in x,y,z & No & High \\
        Fourier transform of displacement in x & No & Low \\
        Wavelet transform of displacement in x & No & Low \\
        Trajectory crossing of displacement in x & \textbf{\textit{Yes}} & Very Low \\
        Slope variations of displacement in x & \textit{\textcolor{black}{No}} & \textit{\textcolor{black}{Very Low}} \\
        \hline
        \end{tabular}
        \caption{Comparison of different encoding strategies for numerical dispersion detection during exploitation (training always requires multiple simulations).}
    \end{table}
\section{Conclusion and Perspectives}
\label{sec:conclusion_perspectives}

The results show that the proposed post-processing tool can identify regions sensitive to numerical dispersion using a limited number of simulations for training and when applied to new computations. The comparative analysis suggests that the RRAE-based framework is more effective than the Random Forest baseline on the studied dataset. Among the tested signal representations, wavelet-based and slope-variation-based inputs appear to be the most promising: wavelet features improve class separation, whereas slope variations provide the highest classification performance.

Its use in an industrial context nevertheless requires further validation to quantify the robustness of its predictions with respect to changes in the \DT{}. More generally, these results highlight the need to assess the temporal validity of the proposed framework under evolving technical definitions.

\paragraph{}

Future work should focus on the following directions:

\begin{itemize}
    \item \textbf{Assessing temporal validity under evolving technical definitions.}  
    Crash models continuously evolve throughout vehicle development through geometric modifications of the body structure, changes in subsystems, material updates, evolutions in assembly methods, and local thickness variations driven by mass optimization. These evolutions may be classified into two categories: \textit{architectural variations}, which affect the overall vehicle layout and may significantly modify mass distribution, stiffness, and crash mechanisms; and \textit{micro-variations}, which correspond to more localized changes that may still influence numerical dispersion. A key challenge is therefore to evaluate how the accuracy of the tool degrades when the \DT{} deviates from the context in which the model was developed.

    \item \textbf{Defining validation strategies for robustness analysis.}  
    Validation should be understood here as a quantitative evaluation of performance drift under changes in the \DT{}, rather than as a simple functional verification. Relevant indicators include classification accuracy, recall for dispersed \neu{}s, the spatial stability of detected regions, and the drift of dispersion scores.

    \item \textbf{Performing longitudinal validation during a vehicle development project.}  
    One possible strategy consists of selecting representative project milestones, repeating simulations at each stage, applying the model built on the initial configuration, and evaluating how performance metrics evolve as technical modifications accumulate. Although such a longitudinal analysis could not be carried out within the temporal scope of this study, it would provide a direct measure of progressive model degradation.

    \item \textbf{Extending validation across distinct vehicle projects.}  
    A second strategy consists of applying a model trained on one vehicle project to another in order to assess robustness under more pronounced variations of the \DT{}. This involves measuring classification performance, comparing the spatial localization of detected zones, and quantifying the associated loss of accuracy. In addition, the coexistence of multiple vehicle projects developed in parallel could be exploited to collect configurations of the \DT{} at regular intervals. Such inter-project sampling would act as a surrogate for longitudinal evolution within a single project and would provide a practical estimate of model stability in an industrial context.

    \item \textbf{Confirming the relevance of slope variations for robustness studies.}  
    To reliably assess performance drift, the chosen indicators should exhibit limited dependence on geometry or mesh representation, so that observed variations primarily reflect changes in the \DT{} rather than representation artifacts. In this respect, slope variations of trajectories appear particularly relevant, as they reduce structural dependence while preserving local signal dynamics. They therefore provide a suitable basis for inter-\DT{} robustness analyses.

    \item \textbf{Extending the framework beyond node-level variability.}  
    Beyond node-level dispersion, variability may also affect the chronological development of the crash response. Variations in the timing or ordering of structural events can alter load redistribution mechanisms and influence the final structural state. Future work should therefore investigate the stability of the crash sequence itself.

    \item \textbf{Linking dispersion detection to physical mechanisms.}  
    The identification of sensitive \neu{}s constitutes a first step toward analyzing the mechanisms that amplify numerical dispersion, such as contact activation, local buckling, or modifications of load paths. Extending the approach toward the evaluation of crash scenario stability would explicitly integrate the temporal dimension of structural events and strengthen the reliability of design decisions.
\end{itemize}

\newpage
\bibliographystyle{plainnat}
\bibliography{bibliography}
\end{document}